\def\year{2020}\relax
\documentclass[letterpaper]{article} 
\usepackage{aaai20}  
\usepackage{times}  
\usepackage{helvet} 
\usepackage{courier}  
\usepackage[hyphens]{url}  
\usepackage{graphicx} 
\usepackage{graphicx}  
\usepackage{booktabs}
\usepackage{amsfonts}
\usepackage{amsmath}
\usepackage{microtype}
\usepackage[binary-units]{siunitx}
\title{LCD: Learned Cross-Domain Descriptors for 2D-3D Matching}

\author{
  Quang-Hieu Pham\textsuperscript{\rm 1}\hspace{0.2in}
  Mikaela Angelina Uy\textsuperscript{\rm 2}\hspace{0.2in}
  Binh-Son Hua\textsuperscript{\rm 3}\hspace{0.2in}
  Duc Thanh Nguyen\textsuperscript{\rm 4}
  \\
  \bf \Large Gemma Roig\textsuperscript{\rm 5}\hspace{0.2in}
  Sai-Kit Yeung\textsuperscript{\rm 6}
  \\
  \textsuperscript{\rm 1}Singapore University of Technology and Design\hspace{0.2in}
  \textsuperscript{\rm 2}Stanford University\hspace{0.2in}
  \textsuperscript{\rm 3}The University of Tokyo
  \\
  \textsuperscript{\rm 4}Deakin University\hspace{0.2in}
  \textsuperscript{\rm 5}Geothe University of Frankfrut am Main\hspace{0.2in}
  \textsuperscript{\rm 6}Hong Kong University of Science and Technology
}

\newcommand{\ie}{\textit{i}.\textit{e}.}
\newcommand{\eg}{\textit{e}.\textit{g}.}

\newcommand{\K}{\mathcal{K}}
\newcommand{\Image}{I}
\newcommand{\Point}{P}
\newcommand{\Loss}{\mathcal{L}}
\newcommand{\norm}[1]{\left\lVert#1\right\rVert}
\newcommand{\Real}{\mathbb{R}}

\graphicspath{{./figures/}}

\begin{document}

\maketitle

\begin{abstract}
  In this work, we present a novel method to learn a \emph{local cross-domain descriptor} for 2D image and 3D point cloud matching.
  Our proposed method is a dual auto-encoder neural network that maps 2D and 3D input into a shared latent space representation.
  We show that such local cross-domain descriptors in the shared embedding are more discriminative than those obtained from individual training in 2D and 3D domains.
  To facilitate the training process, we built a new dataset by collecting $\approx 1.4$ millions of 2D-3D correspondences with various lighting conditions and settings from publicly available RGB-D scenes.
  Our descriptor is evaluated in three main experiments: 2D-3D matching, cross-domain retrieval, and sparse-to-dense depth estimation.
  Experimental results confirm the robustness of our approach as well as its competitive performance not only in solving cross-domain tasks but also in being able to generalize to solve sole 2D and 3D tasks.
  Our dataset and code are released publicly at \url{https://hkust-vgd.github.io/lcd}.
\end{abstract}

\section{Introduction}
\label{sec:intro}

Computer vision tasks such as structure-from-motion, visual content retrieval require robust descriptors from both 2D and 3D domains. Such descriptors, in their own domain, can be constructed from low-level features, \eg, colors, edges, etc.
In image matching, a well-known task in computer vision, several hand-crafted local descriptors, \eg, SIFT~\cite{lowe2004sift}, SURF~\cite{bay2006surf} have been proposed.
With the advent of deep learning, many robust 2D descriptors are learned automatically using deep neural networks~\cite{simo2015deepdesc,kumar2016patchmatch}.
These learned descriptors have shown their robustness and advantages over the hand-crafted counterparts.
The same phenomenon can also be observed in 3D domain.
For example, hand-crafted 3D descriptors, \eg, FPFH~\cite{rusu2009fpfh}, SHOT~\cite{tombari2010shot}, as well as deep learning based descriptors~\cite{zeng20173dmatch} have been used in many 3D tasks, such as 3D registrations~\cite{choi2015redwood,zhou2016fgr} and structure-from-motion~\cite{hartley2003multiview}.

While 2D and 3D descriptors are widely available, finding the association between these representations is a challenging task.
There also lacks a descriptor that can capture features in both domains and tailored for cross-domain tasks, for example, 2D to 3D content retrieval.
In general, there is a large discrepancy between 2D and 3D representations.
Data in 2D, \ie, images, can simply be represented by regular grids.
Meanwhile, 3D data can be represented by either meshes, volumes, or point clouds and obtained via an image formation model that is governed by laws of physics and optics.
Even with the recent advent of deep learning, these issues still remain the same: features learned on 2D domain may not be applicable in 3D space and vice versa.

In this work, we attempt to bridge the gap between 2D and 3D descriptors by proposing a novel approach to learn a cross-domain descriptor that works on both 2D and 3D domains.
In particular, we make the following contributions:

\begin{itemize}
\item A novel learned cross-domain descriptor (LCD) that is learned using a dual auto-encoder architecture and a triplet loss.
  Our setup enforces the 2D and 3D auto-encoders to learn a cross-domain descriptor in a shared latent space representation.
  This shared latent space not only provides a common space for 2D-3D matching, but also improves the descriptor performance in single-domain settings.
\item A new public dataset of $\approx 1.4$ millions of 2D-3D correspondences for training and evaluating the cross-domain descriptors matching.
  We built our dataset based on SceneNN~\cite{hua2016scenenn} and 3DMatch~\cite{zeng20173dmatch}.
\item Applications to verify the robustness of our cross-domain descriptor.
  Specifically, we apply the descriptor to solve a sole 2D (image matching) and a sole 3D task (3D registration), and then to a 2D-3D content retrieval task (2D-3D place recognition).
  Experimental results show that our descriptor gives comparable performance to other state-of-the-art methods in all the tasks even when it is not purposely tailored to such particular tasks.
\end{itemize}

\section{Related work}
\label{sec:related}

Local descriptors are crucial components in many applications such as localization~\cite{sattler2017localization}, registration~\cite{choi2015redwood,zhou2016fgr}, Structure-from-Motion (SfM)~\cite{hartley2003multiview}, Simultaneous Localization and Mapping (SLAM)~\cite{durrant2006slam}, and pose estimation~\cite{haralick1991pose}.
In general, 2D descriptors are obtained from the 2D local patches of images, whereas 3D descriptors are often computed from 3D point clouds.

\paragraph{2D descriptors.}
Image descriptors, both fully handcrafted~\cite{lowe2004sift,bay2006surf,rublee2011orb} and partially learned~\cite{brown2010learndesc,tola2009daisy}, have been well studied in early days of computer vision.
Recently, deep learning has been applied for end-to-end learning of 2D descriptors~\cite{chopra2005reid}.
The robustness of the learned descriptors over the handcrafted ones have been proven clearly in image matching.
For example, \cite{zagoruyko2015deepcompare} and \cite{han2015matchnet} proposed a Siamese architecture to learn a similarity score between a given pair of image patches.
However, these methods are computationally expensive as the image patches need to be pairwise passed into the network.
To make the solution tractable, in~\cite{simo2015deepdesc,tian2017l2net}, the descriptor was learned with the same Siamese architecture but matched using Euclidean distance.
This allows the learned descriptor to be used as a direct replacement to traditional hand-crafted descriptors, and nearest neighbor queries could be done efficiently in matching.
Our work is built upon this idea, but instead we learn a cross-domain 2D-3D descriptor.

More recently, triplet networks~\cite{balntas2017hpatches} taking three image patches as input for learning descriptors have been introduced.
These networks showed that learning with a triplet loss~\cite{schroff2015facenet,hermans2017triplet} resulted in a better embedding space.
Further improvements to the triplet loss were studied in~\cite{mishchuk2017hardnet,keller2018scale}.
Joint learning of feature detector and descriptor was explored by~\cite{yi2016lift}.
In general, all of these works take image patches as input and learn a feature space for 2D descriptors.
In contrast, our work aims to learn a shared latent space for both 2D and 3D descriptors.
In addition to leveraging metric learning, we also utilize auto-encoders to learn a more discriminative space.

\paragraph{3D descriptors.}
Unlike 2D descriptors, 3D descriptors for point clouds such as PFH~\cite{rusu2008pfh}, FPFH~\cite{rusu2009fpfh}, and SHOT~\cite{tombari2010shot} do not reach the same level of robustness and maturity.
These methods either require stable surfaces or sufficient point densities.
Deep learning solutions have also been proposed to tackle these problems.
For example, 3DMatch~\cite{zeng20173dmatch} used voxelized patches to calculate 3D local descriptors in convolutions to register RGB-D scans.
\cite{dewan2018lidar} also applied convolutions on 3D voxels to learn local descriptors for LiDAR scans.
However, such architectures cannot be used directly on point cloud due to their irregular input structure that disables convolutions.
To address this issue, \cite{khoury2017cgf} proposed to reduce hand-crafted point cloud descriptor dimension through deep learning for efficient matching.

Recently, PointNet~\cite{qi2017pointnet} introduced the first deep neural network that can directly operate on point clouds.
This network then became the backbone for multiple point-based networks~\cite{deng2018ppfnet,deng2018ppffoldnet,yew20183dfeatnet}.
In particular, PPFNet~\cite{deng2018ppfnet} used point pair features to learn local point cloud descriptors for registration.
3DFeat-Net~\cite{yew20183dfeatnet} proposed a weakly-supervised approach to learn local descriptors with only GPS/INS tags for outdoor data.
PPF-FoldNet~\cite{deng2018ppffoldnet} made use of an auto-encoder to learn point cloud descriptors in an unsupervised manner.
Other deep learning descriptors on point cloud include KeypointNet~\cite{suwajanakorn2018keypointnet}, USIP~\cite{li2019usip}, and PointNetVLAD~\cite{uy2018pointnetvlad}.
These methods address the problem of 3D keypoint detection and LiDAR-based place recognition, and thus, are designed to match 3D structures only. On the other hand, our work handles both 2D image patches and 3D point clouds in a unified manner.

\paragraph{2D-3D cross-domain descriptors.}
\cite{li2015joint} proposed a joint global embedding of 3D shapes and images to solve the retrieval task.
The 3D embeddings were first hand-craftedly constructed and image embeddings were learned to adhere to the 3D embeddings.
In contrast, our network jointly learns both 2D and 3D embeddings for local descriptors.
\cite{xing20183dtnet} proposed a network, called 3DTNet, that receives both 2D and 3D local patches as input.
However, 3DTNet was only designed for 3D matching and the network used 2D features as auxiliary information to make 3D features more discriminative.
Some other works also established the connection between 2D and 3D for specific applications such as object pose estimation~\cite{lim2013ikea,xiao2012cuboid} and 3D shape estimation~\cite{hejrati2012analyzing}.
Recently, \cite{feng20192d3dmatchnet} proposed a deep network to match 2D and 3D patches with a triplet loss for outdoor localization.
Our work differs from this method in that our goal is not to learn an application-specific descriptor, but a cross-domain descriptor for generalized 2D-3D matching that can be used in various tasks as proven in our experiments.

\section{Learned cross-domain descriptors (LCD)}
\label{sec:method}

\begin{figure*}[t]
  \centering
  \includegraphics[width=\linewidth]{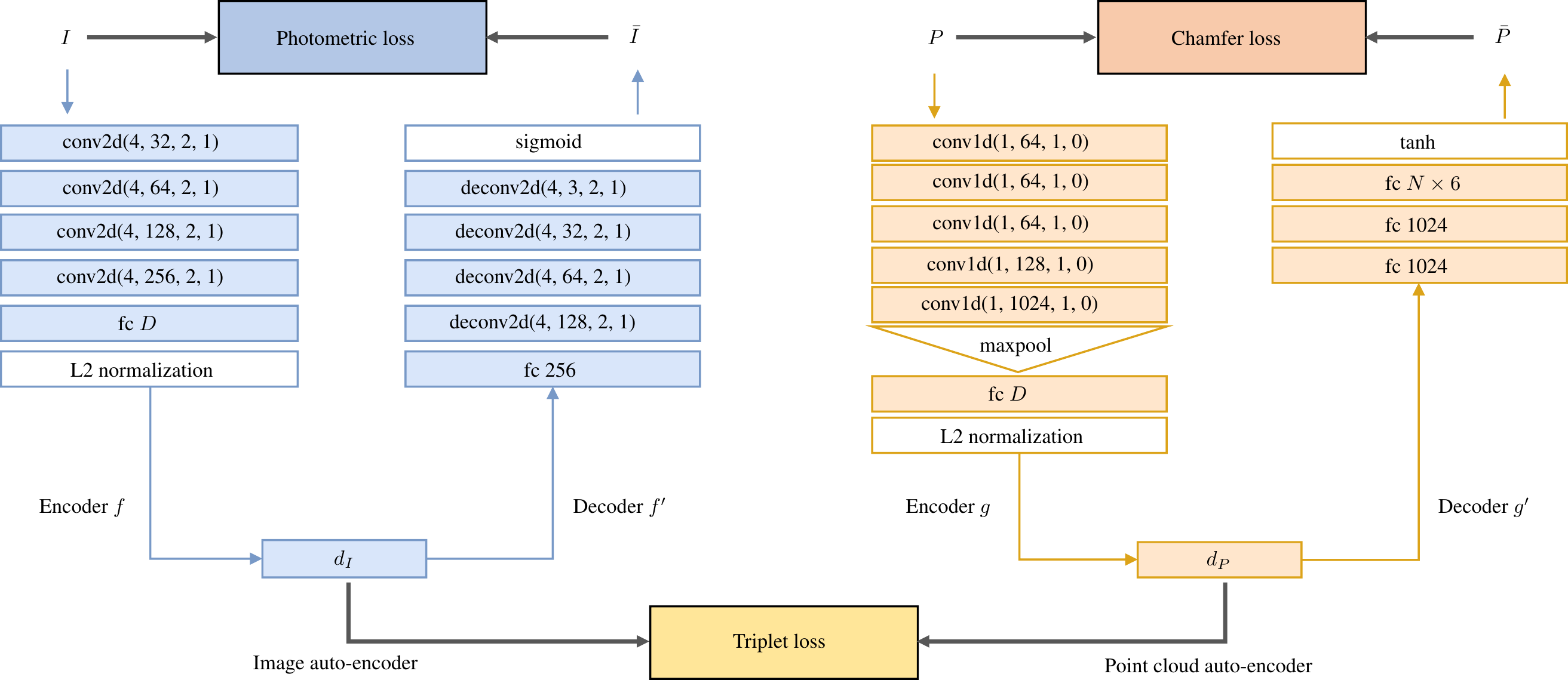}
  \caption{
    \textbf{Our proposed network consists of a 2D auto-encoder and a 3D auto-encoder.}
    The input image and point cloud data is reconstructed with a photometric and a Chamfer loss, respectively.
    The reconstruction losses ensures features in the embedding to be discriminative and representative.
    The similarity between the 2D embedding $d_I$ and the 3D embedding $d_P$ is further regularized by a triplet loss.
    Diagram notation: \texttt{fc} for fully-connected, \texttt{conv/deconv(kernel\_size, out\_dim, stride, padding)} for convolution and deconvolution, respectively.
    Each convolution and deconvolution is followed by a ReLU activation and a batch normalization by default.
  }
  \label{fig:p2pnet}
\end{figure*}

\subsection{Problem definition}
Let $\Image \in \Real^{W \times H \times 3}$ be a colored image patch of size $W \times H$ and represented in conventional RGB color space.
Similarly, let $\Point \in \Real^{N \times 6}$ be a colored point cloud of $N$ points, each point is represented by its coordinates $(x,y,z) \in \Real^3$ and RGB color.

Our goal of learning a cross-domain descriptor is to find two mappings $f: \Real^{W \times H\times 3} \mapsto \mathcal{D}$ and $g: \Real^{N \times 6} \mapsto \mathcal{D}$ that maps the 2D and 3D data space to a shared latent space $\mathcal{D} \subseteq \Real^D$, where $D$ is the dimension of the embedding such that for each pair of 2D-3D correspondence $(\Image, \Point)$, their mappings are as similar as possible.
Mathematically, given a distance function $\mathcal{F}$ and two descriptors $d_\Image, d_\Point \in \mathcal{D}$, if $\Image$ and $\Point$ are represented the same underlying geometry, then $\mathcal{F}(d_\Image, d_\Point) < m$, where $m$ is a predefined margin.

In addition to mapping data to descriptors, we also aim to learn the inverse mapping functions $f': \mathcal{D} \mapsto \Real^{W \times H\times 3}$ and $g': \mathcal{D} \mapsto \Real^{N \times 6}$.
Being able to reconstruct data from descriptors, these inverse mappings are beneficial in downstream applications such as 3D sparse-to-dense depth estimation, as shown later in our experiments.

\subsection{Network architecture}
Inspired by the success of using auto-encoders in construction of descriptors~\cite{deng2018ppffoldnet}, we propose a novel dual auto-encoder architecture to learn descriptors.
Our model is a two-branch network architecture, where one branch encodes 3D features, and the other branch encodes 2D features.
The two branches are then jointly optimized using a triplet loss enforcing the similarity of embeddings generated by the two branches as well as the 2D/3D reconstruction losses.
Our network architecture is illustrated in Figure~\ref{fig:p2pnet}.

For the 2D branch, our 2D auto-encoder takes input a colored image patch of size $64 \times 64$ and processes it through a series of convolutions with ReLU activation in order to extract image features.
For the 2D decoder, we use a series of transpose convolutions with ReLU to reconstruct the image patch.
For the 3D branch, we adopt the well-known PointNet architecture~\cite{qi2017pointnet}, employing a series of fully-connected layers then max-pooling to calculate a global feature.
To reconstruct the colored point cloud, we employ another series of fully-connected layers which outputs a colored point cloud of size $N \times 6$.
To enforce a shared representation, the two auto-encoders are tied up between their bottlenecks by optimizing a triplet loss. The final training loss combines photometric loss, Chamfer loss, and triplet loss as follows.

\paragraph{Photometric loss.}
The 2D auto-encoder loss is defined by the photometric loss, which is the mean squared error between the input 2D patch $\Image$ and the reconstructed patch $\bar{\Image}$:
\begin{equation}
  \label{eqn:mse}
  \Loss_{mse} = \frac{1}{W \times H}
  \sum_{i=1}^{W \times H}\norm{\Image_i - \bar{\Image}_i}^2,
\end{equation}
where $\Image_i$ and $\bar{\Image}_i$ denote the $i^{th}$ pixel in the input and reconstructed image patches, respectively.

\paragraph{Chamfer loss.}
To optimize the 3D auto-encoder network, we need to compute the distance between the input point set $\Point$ and the reconstructed point set $\bar{\Point}$.
We measure this distance via the well known Chamfer distance:
\begin{align}
  \label{eqn:chamfer}
  \Loss_{chamfer} = \max \biggl\{
  &\frac{1}{|\Point|}\sum_{p \in \Point}\min_{q \in \bar{\Point}}\norm{p - q}_2, \notag \\
  &\frac{1}{|\bar{\Point}|} \sum_{q \in \bar{\Point}}\min_{p \in \Point}\norm{p - q}_2
  \biggr\}.
\end{align}

\paragraph{Triplet loss.}
To enforce the similarity in the embeddings generated by the 2D and 3D branch, \ie, a 2D image patch and its corresponding 3D structures should have similar embeddings, we employ the triplet loss function.
This loss minimizes the distance between an anchor and a positive, while maximizes the distance between the anchor and a negative.
Following~\cite{hermans2017triplet}, we perform online batch-hardest negative mining that can improve both train and test performance.
The triplet loss function can be written as follows:
\begin{equation}
  \label{eqn:triplet}
  \Loss_{triplet} = \max(\mathcal{F}(d_a, d_p) - \mathcal{F}(d_a, d_n) + m, 0),
\end{equation}
where $m$ is the margin and $\mathcal{F}$ is the distance function. $(d_a, d_p, d_n)$ is a triplet consisting of an anchor, a positive, and a hardest negative, respectively.

\paragraph{Training loss.}
In summary, the loss function that is used to train our network is defined as:
\begin{equation}
  \label{eqn:loss}
  \Loss = \alpha \cdot \Loss_{mse} + \beta \cdot \Loss_{chamfer} + \gamma \cdot \Loss_{triplet},
\end{equation}
where $\alpha$, $\beta$, and $\gamma$ are the weights to emphasis the importance of each sub-network in the training process.
We set $\alpha = \beta = \gamma = 1$ in our implementation.

The proposed architecture holds several advantages.
First, the 2D and 3D branch capture important features in 2D and 3D domains.
When these branches are trained jointly, domain invariant features would be learned and integrated into the embeddings.
Second, having auto-encoders in the architecture enables the transformation of descriptors across 2D and 3D domains as shown in our experiments.

\subsection{Implementation details}
Our network requires a dataset of 2D-3D correspondences to train.
To the best of our knowledge, there is no such publicly available dataset.
Therefore, we build a new dataset of 2D-3D correspondences by leveraging the availability of several 3D datasets from RGB-D scans.
In this work, we use the data from SceneNN~\cite{hua2016scenenn} and 3DMatch~\cite{zeng20173dmatch}.
SceneNN is a comprehensive indoor dataset scanned by handheld RGB-D sensor with fine-grained annotations.
3DMatch dataset is a collection of existing RGB-D scenes from different works~\cite{glocker2013reloc,xiao2013sun3d,valentin2016learning,dai2017bundlefusion,henry2013patchvolumes,halber2017fine2coarse}.
We follow the same train and test splits from~\cite{zeng20173dmatch} and~\cite{hua2018pointwise}.
Our training dataset consists of $110$ RGB-D scans, of which $56$ scenes are from SceneNN and $54$ scenes are from 3DMatch.
The models presented in our experiments are all trained on the same dataset.

The 2D-3D correspondence data is generated as follows.
Given a 3D point which is randomly sampled from a 3D point cloud, we extract a set of 3D patches from different scanning views.
To find a 2D-3D correspondence, for each 3D patch, we re-project its 3D position into all RGB-D frames for which the point lies in the camera frustum, taking occlusion into account.
We then extract the corresponding local 2D patches around the re-projected point.
In total, we collected $1,465,082$ 2D-3D correspondences, with varying lighting conditions and settings.

Our network is implemented in PyTorch.
The network is trained using SGD optimizer, with learning rate set to $0.01$.
We train our network on a cluster equipped with NVIDIA V100 GPUs and $\SI{256}{\giga\byte}$ of memory.
It takes around $17$ hours to train our network, stopping after $250$ epochs.

\section{Experiments}
\label{sec:experiments}

In this section, we evaluate our proposed cross-domain descriptor under a wide range of applications, showing that the learned descriptor can work on both 2D and 3D domains.
We also explore the effect of the output feature dimension $D$ on the descriptor's performance.
In our experiments, we train and test with $D \in \{64, 128, 256\}$, denoted as \emph{LCD-D*} in the results.
We evaluate the performance of our 2D descriptor on the task of image matching.
Then, we demonstrate the capability of our 3D descriptor in the global registration problem.
Our cross-domain descriptor also enables unique applications, such as 2D-3D place recognition and sparse-to-dense depth estimation.
Finally, we validate our network design by conducting in-depth ablation study.

\subsection{2D image matching}
\label{sec:matching}
We first evaluate our descriptor in a classic 2D computer vision task --- image matching.
We use the SceneNN dataset, which contains around $100K$ of RGB-D frames with ground-truth pose estimation.
We use $20$ scenes for testing, following the same split from~\cite{hua2018pointwise}.
We consider the following descriptors as our competitors: traditional hand-crafted descriptors \emph{SIFT}~\cite{lowe2004sift}, \emph{SURF}~\cite{bay2006surf}, and a learned descriptor \emph{SuperPoint}~\cite{detone2018superpoint}.
We also show the result from \emph{PatchNetAE}, which is the single 2D branch of our network trained only with image patches.

\begin{figure}[t]
  \centering
  \includegraphics[width=\linewidth]{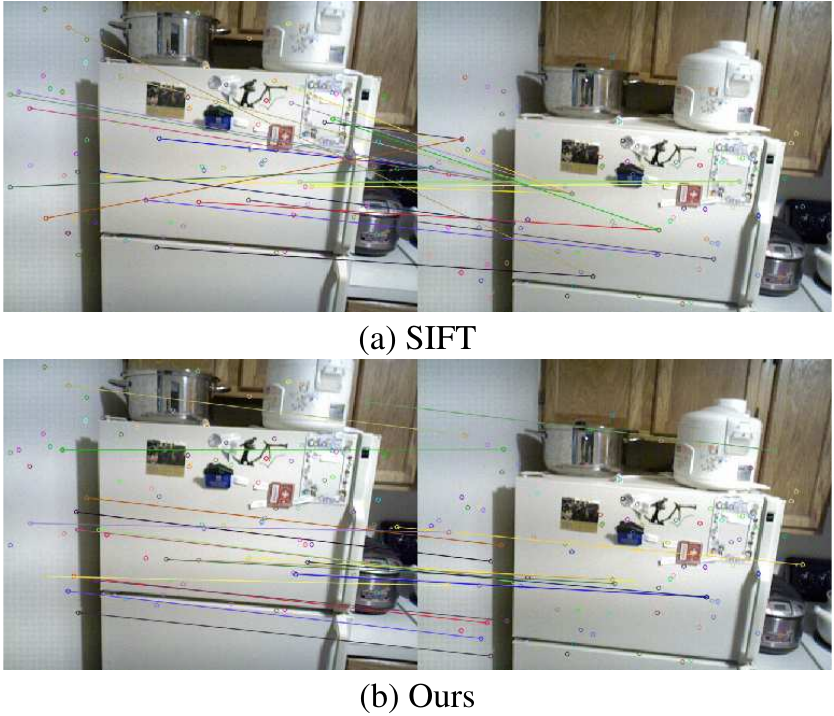}
  \caption{
    \textbf{Qualitative 2D matching comparison between SIFT and our proposed descriptor.}
    Our descriptor can correctly identify features from the wall and the refrigerator, while SIFT~\cite{lowe2004sift} fails to differentiate them.
  }
  \label{fig:2dmatch}
\end{figure}

To evaluate the performance over different baselines, we sample image pairs at different frame difference values: $10$, $20$, and $30$.
SuperPoint is an end-to-end keypoint detector and descriptor network.
Since we are only interested in the performance of descriptors, to give a fair comparison, we use the same keypoints extracted by SuperPoint for all of the descriptors.
Matching between two images is done by finding the nearest descriptor.
We use precision as our evaluation metric, which is the number of true matches over the number of predicted matches.
Table~\ref{tab:2dmatch} shows the performance of our descriptor compared to other methods.
Overall, our method outperforms other traditional handcrafted descriptors by a margin, and gives a favorable performance compared to other learning-based method, \ie, SuperPoint.
An example of 2D matching visualization is provided in Figure~\ref{fig:2dmatch}.
As can be seen, our descriptor gives a stronger performance with more correct matches.

\begin{table*}[t]
  \centering
  \caption{
    \textbf{2D matching results (precision) on the SceneNN dataset.}
    Best results are marked in bold.
  }
  \label{tab:2dmatch}
  \begin{tabular}{lrrrrrrr}
    \toprule
    Frame difference & SIFT  & SURF  & SuperPoint     & PatchNetAE & LCD-D256       & LCD-D128 & LCD-D64 \\
    \midrule
    10               & 0.252 & 0.231 & 0.612          & 0.613      & \textbf{0.625} & 0.604    & 0.591   \\
    20               & 0.183 & 0.157 & \textbf{0.379} & 0.360      & 0.373          & 0.364    & 0.347   \\
    30               & 0.125 & 0.098 & 0.266          & 0.245      & \textbf{0.267} & 0.256    & 0.239   \\
    \midrule
    Average          & 0.187 & 0.162 & 0.419          & 0.406      & \textbf{0.422} & 0.408    & 0.392   \\
    \bottomrule
  \end{tabular}
\end{table*}

\begin{table*}[t]
  \centering
  \caption{
    \textbf{3D registration results (recall) on the 3DMatch benchmark.}
    Best results are marked in bold.
  }
  \label{tab:3dreg}
  \begin{tabular}{lrrrrrrrr}
    \toprule
            & CZK   & FGR   & 3DMatch & 3DSmoothNet    & PointNetAE & LCD-D256       & LCD-D128 & LCD-D64        \\
    \midrule
    Kitchen & 0.499 & 0.305 & 0.853   & 0.871          & 0.766      & \textbf{0.891} & 0.889    & 0.891          \\
    Home 1  & 0.632 & 0.434 & 0.783   & \textbf{0.896} & 0.726      & 0.783          & 0.802    & 0.757          \\
    Home 2  & 0.403 & 0.283 & 0.610   & \textbf{0.723} & 0.579      & 0.629          & 0.616    & 0.610          \\
    Hotel 1 & 0.643 & 0.401 & 0.786   & 0.791          & 0.786      & 0.808          & 0.813    & \textbf{0.841} \\
    Hotel 2 & 0.667 & 0.436 & 0.590   & \textbf{0.846} & 0.680      & 0.769          & 0.821    & 0.821          \\
    Hotel 3 & 0.577 & 0.385 & 0.577   & \textbf{0.731} & 0.731      & 0.654          & 0.654    & 0.692          \\
    Study   & 0.547 & 0.291 & 0.633   & 0.556          & 0.641      & \textbf{0.662} & 0.628    & 0.650          \\
    MIT Lab & 0.378 & 0.200 & 0.511   & 0.467          & 0.511      & \textbf{0.600} & 0.533    & 0.578          \\
    \midrule
    Average & 0.543 & 0.342 & 0.668   & \textbf{0.735} & 0.677      & 0.725          & 0.720    & 0.730          \\
    \bottomrule
  \end{tabular}
\end{table*}

\subsection{3D global registration}
\label{sec:registration}
To demonstrate a practical use of our descriptor, we combine it with RANSAC for the 3D global registration task.
Given two 3D fragments from scanning, we uniformly downsample the fragments to obtain the keypoints.
For every interest point, we form a local patch by taking points within a neighborhood of $\SI{30}{\centi\meter}$.
The 3D descriptors are then computed for all of these keypoints.
We match the two sets of keypoints with nearest neighbor search and use RANSAC to estimate the final rigid transformation.

We use the 3DMatch Benchmark~\cite{zeng20173dmatch} to evaluate the 3D matching performance of our descriptor, which contains $8$ scenes for testing.
3DMatch already provided test fragments fused from consecutive depth frames.
However, these fragments lack color information, which is required for our descriptor, so we modified the pipeline to generate another version with color.

We follow the same evaluation process introduced by~\cite{choi2015redwood}, using recall as the evaluation metric.
Given two non-consecutive scene fragments $P_i$ and $P_j$, the predicted relative rigid transformation $T_{ij}$ is a true positive if (1) over $30\%$ of $T_{ij}P_i$ overlaps with $P_j$ and (2) $T_{ij}$ is sufficiently close to the ground-truth transformation $\hat{T}_{ij}$.
Specifically, $T_{ij}$ is correct if the RMSE of ground-truth correspondences $\hat{\K}_{ij}$ is below a threshold $\tau = \SI{0.2}{\meter}$:
\begin{equation}
  \frac{1}{|\hat{\K}_{ij}|}\sum_{(\hat{p}, \hat{q}) \in \hat{\K}_{ij}}\norm{T_{ij}\hat{p} - \hat{q}}^2 < \tau^2.
\end{equation}

\begin{figure*}[t]
  \centering
  \includegraphics[width=0.9\linewidth]{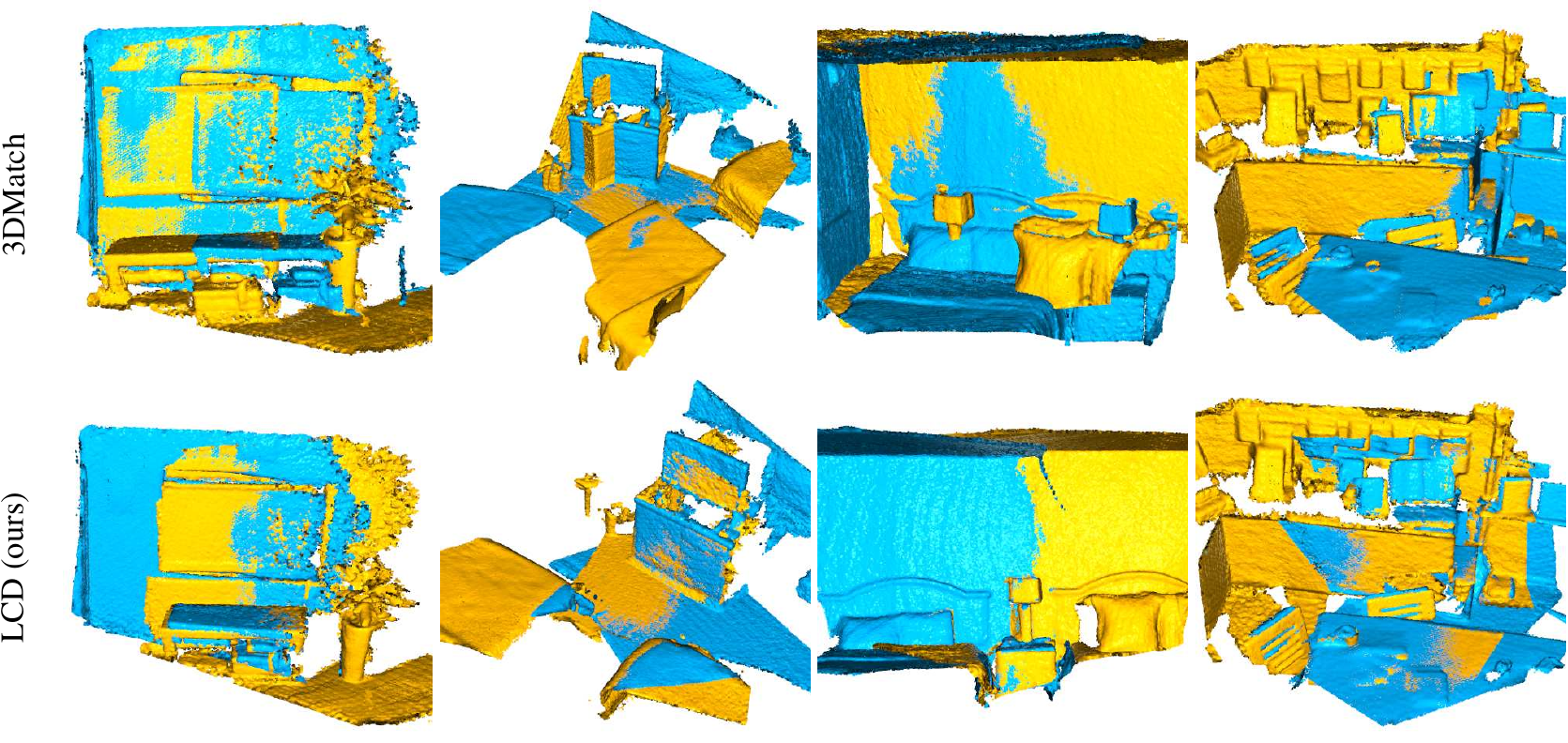}
  \caption{
    \textbf{Qualitative results on the 3DMatch benchmark.}
    Our method is able to successfully align pair of fragments in different challenging scenarios by matching local 3D descriptors, while 3DMatch~\cite{zeng20173dmatch} fails in cases when there are ambiguities in geometry.
  }
  \label{fig:registration}
\end{figure*}

Table~\ref{tab:3dreg} lists the recall of different algorithms on the 3DMatch benchmark.
\emph{CZK}~\cite{choi2015redwood} uses FPFH descriptor~\cite{rusu2009fpfh} with RANSAC to prune false matches.
\emph{3DMatch}~\cite{zeng20173dmatch} employs the same RANSAC-based pipeline, using their own voxel-based learned descriptor.
\emph{FGR}~\cite{zhou2016fgr} is a fast global registration algorithm which does not rely on iterative sampling.
\emph{3DSmoothNet}~\cite{gojcic20193dsmoothnet} is a newly proposed method that uses a voxelized smooth density value representation.
\emph{PointNetAE} is just the single 3D branch of our network trained only with the 3D data in the dataset.
Overall, our descriptor with RANSAC outperforms others by a significant margin. We also show additional qualitative results in Figure~\ref{fig:registration}.

\subsection{2D-3D place recognition}
\label{sec:retrieval}

\begin{figure}[t]
  \centering
  \includegraphics[width=0.75\linewidth]{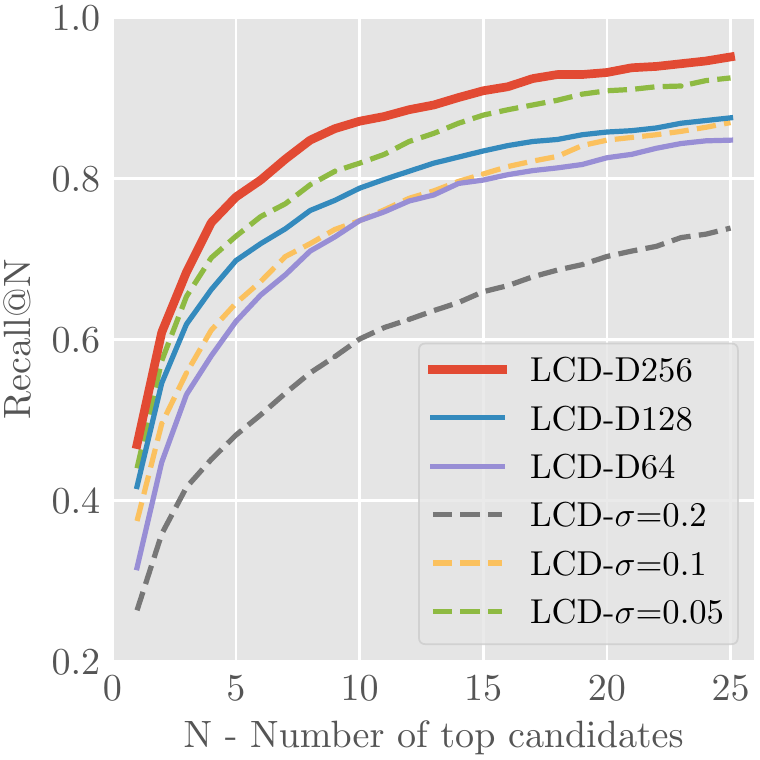}
  \caption{
    \textbf{Results of the 2D-3D place recognition task.}
    \emph{LCD-D256}, \emph{LCD-D128}, and \emph{LCD-D64} indicate descriptor with different dimensions.
    While being effective, our cross-domain descriptor also demonstrates the robustness to input noise, with $\textit{LCD-}\sigma$ indicating the results when adding Gaussian noise with standard deviation $\sigma$ into the query images.
  }
  \label{fig:recall}
\end{figure}

We further evaluated our local cross-domain descriptor with 2D-to-3D place recognition.
Unlike previous works in single-domain place recognition~\cite{torii2015densevlad,arandjelovic2016netvlad}, our task is to find the corresponding 3D geometry submap in the database given a query 2D image.
We only assume that raw geometries are given, without additional associated image descriptors and/or camera information as often seen in the camera localization problem~\cite{zeisl2015cpv,sattler2015hyperpoints}.
With increasing availability of 3D data, such 2D-3D place recognition becomes practical as it allows using Internet photos to localize a place in 3D.
To the best of our knowledge, there has been no previous report about solving this cross-domain problem.

Here we again use the SceneNN dataset~\cite{hua2016scenenn}.
Following the split from~\cite{hua2018pointwise}, we use $20$ scenes for evaluation.
To generate geometry submaps, we integrate every $100$ consecutive RGB-D frames.
In total, our database is consisted of $1,191$ submaps from various lighting conditions and settings such as office, kitchen, bedroom, etc.
The query images are taken directly from the RGB frames, such that every submap has at least one associated image.

We cast this 2D-3D place recognition problem as an retrieval task.
Inspired by the approach from Dense\-VLAD~\cite{torii2015densevlad}, for each 3D submap, we sample descriptors on a regular voxel grid.
These descriptors are then aggregate into a single compact VLAD descriptor~\cite{jegou2010vlad}, using a dictionary of size $64$.
To extract the descriptor for an image, we follow the same process, but on a 2D grid.
The query descriptor are matched with the database to retrieve the final results.

We follow the standard place recognition evaluation procedure~\cite{torii2015densevlad,arandjelovic2016netvlad}.
The query image is deemed to be correctly localized if at least one of the top $N$ retrieved database submaps is within $d = \SI{0.5}{\meter}$ and $\theta = 30^{\circ}$ from the ground-truth pose of the query.

\begin{figure}[t]
  \centering
  \includegraphics[width=\linewidth]{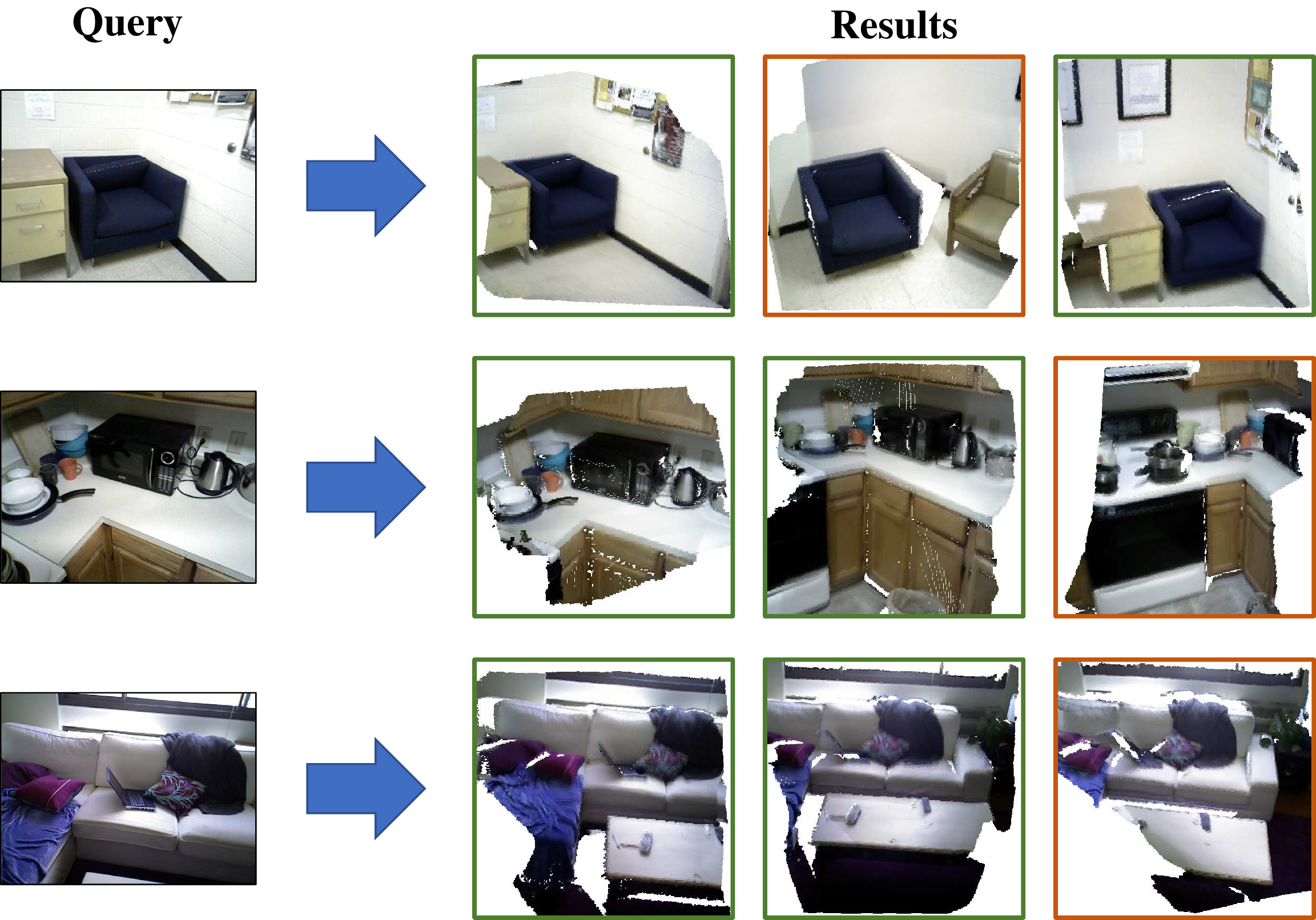}
  \caption{
    \textbf{Top-3 retrieval results of the 2D-3D place recognition task using our descriptor.}
    Green/red borders mark correct/incorrect retrieval.
    Best view in color.
  }
  \label{fig:retrieval}
\end{figure}

We plot the fraction of correct queries (recall@N) for different value of $N$, as shown in Figure~\ref{fig:recall}.
Representative top-3 retrieval results are shown in Figure~\ref{fig:retrieval}.
It can be seen that our local cross-domain descriptor are highly effective in this 2D-to-3D retrieval task.

\begin{figure}[t]
  \centering
  \includegraphics[width=\linewidth]{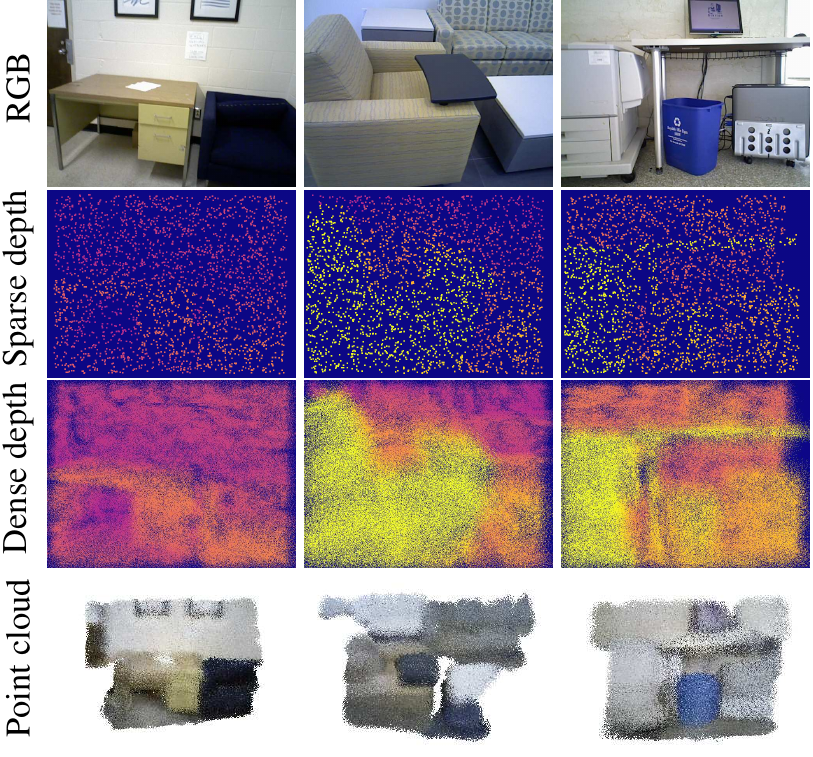}
  \caption{
    \textbf{Sparse-to-dense depth estimation results.}
    Inputs are a RGB image and $2048$ sparse depth samples.
    Our network estimates dense depth map by reconstructing local 3D points.
    Best view in color.
  }
  \label{fig:depth}
\end{figure}

\subsection{Sparse-to-dense depth estimation}
\label{sec:depth}
As we have the inverse mapping from the shared latent space back to 2D and 3D due to the use of auto-encoders, we can apply the local cross-domain descriptors for dense depth estimation.
Particularly, here we show how to enrich depth from a color image with local predictions from the color channel.
We perform sparse-to-dense depth estimation, where dense depth is predicted from sparse samples~\cite{mal2018std}.
This can be beneficial in robotics, augmented reality, or 3D mapping, where the resolution of depth sensor is limited but the resolution of the color sensor is very high.

Given an RGB-D image, we first take the RGB and sample uniform 2D patches on a $50 \times 50$ regular grid.
From these 2D patches, we first encode them using the 2D encoder, and decode using the 3D decoder to reconstruct the local point clouds.
We assemble these local point clouds into one coherent point cloud by using the input depth samples.
Finally, the dense depth prediction is calculated by projecting the dense 3D point cloud back to the image plane.
Figure~\ref{fig:depth} shows some qualitative results from cross-domain network.

Here we also compare our method against \emph{FCRN}~\cite{laina2016fcrn} on the SceneNN dataset.
Note that FCRN only uses RGB images as input, instead of RGB and sparse depth.
We use the same $1,191$ RGB-D frames in the 2D-3D place recognition experiment for this evaluation, keeping the same training configuration.
Evaluation metrics include: absolute mean relative error (REL), root-mean-square error (RMSE), and percentage of pixels within a threshold ($\delta_i$).
We report the evaluation results in Table~\ref{tab:depth}.

\begin{table}[t]
  \centering
  \caption{
    \textbf{Quantitative results of indoor depth estimation on the SceneNN dataset.}
    Our method outperforms conventional depth estimation method FCRN~\cite{laina2016fcrn} by a margin.
    Note that our method uses both RGB image and sparse depth samples, while FCRN only use RGB image as input.
    Best results are marked in bold.
  }
  \label{tab:depth}
  \begin{tabular}{lrrrrr}
    \toprule
             & REL            & RMSE           & $\delta_1$     & $\delta_2$     & $\delta_3$     \\
    \midrule
    FCRN     & 0.458          & 0.548          & 0.467          & 0.791          & \textbf{0.935} \\
    LCD-D64  & \textbf{0.187} & \textbf{0.446} & \textbf{0.861} & \textbf{0.894} & 0.908          \\
    LCD-D128 & 0.193          & 0.458          & 0.857          & 0.890          & 0.903          \\
    LCD-D256 & 0.194          & 0.459          & 0.858          & 0.890          & 0.903          \\
    \bottomrule
  \end{tabular}
\end{table}

\subsection{Ablation study}
\label{sec:ablation}

\paragraph{Robustness to noise.}
Since our descriptor uses color information in the training process, there is a risk that it might be over-fitting to color information, and not robust to the lighting changes in either 2D or 3D domain.
To demonstrate the robustness of our method, we simulate color changes by adding Gaussian noise of standard deviation $\sigma$ to the input image patches.
We then run the 2D-3D place recognition experiment again with different levels of noisy input, and compare to the original result.
Figure~\ref{fig:recall} shows the performance of our descriptor under different Gaussian noises, named $\text{LCD-}\sigma$.
With a moderate level of noise ($\sigma=0.1$), our descriptor still get a very good performance, achieving $\approx 75\%$ recall at $10$ top candidates.
This study shows that there is no need for the color in 2D patch and 3D point cloud to be identical, and our proposed descriptor is robust to input noise.

\paragraph{Single-domain vs. cross-domain.}
We also compare our cross-domain descriptor with descriptors trained on single domain.
As shown in Table~\ref{tab:2dmatch} and Table~\ref{tab:3dreg}, we train two single auto-encoder models on either 2D or 3D domain, denoting as \emph{PatchNetAE} and \emph{PointNetAE}, respectively.
Our learned cross-domain descriptor consistently outperforms single-domain models by a margin.
This result implies that not only learning cross-domain descriptors benefits downstream applications, but it also improves the performance on single-domain
tasks.

\paragraph{Efficiency.}
Our proposed descriptor is very efficient and can be used in real-time applications.
Generating descriptors only requires a forward pass on the decoder, which only takes $\SI{3.4}{\milli\second}$ to compute $2048$ descriptors, compared to $\SI{4.5}{\milli\second}$ when using ORB~\cite{rublee2011orb}.
The network only uses around $\SI{1}{\giga\byte}$ of GPU memory for inference.

\section{Conclusion}
\label{sec:conclusion}
We propose a new local cross-domain descriptor that encodes 2D and 3D structures into representations in the same latent space.
Our learned descriptor is built based on auto-encoders that are jointly trained to learn domain invariant features and representative and discriminative features of 2D/3D local structures.
We found that local cross-domain descriptors are more discriminative than single-domain descriptors.
We demonstrated that cross-domain descriptors are effective to 2D-3D matching and retrieval tasks.
While our descriptor is task agnostic, we demonstrated that the descriptors are robust when being used for image matching and 3D registration, achieving competitive results with other state-of-the-art methods.
A potential future work is to integrate our descriptor with keypoint detectors to make full image matching and retrieval becomes possible.

\paragraph{Acknowledgements.}
This research project is partially supported by an internal grant from HKUST (R9429).

\section{Embedding visualization}
We visualize common embeddings (\ie, common representations generated by the 2D and 3D auto-encoders) from our proposed network by mapping the high-dimensional descriptors (a vector of $256$ values) into 2D using t-SNE visualization~\cite{maaten2008tsne}.
Particularly, we pass each 2D image patch through our network and collect its descriptor and determine its 2D location using t-SNE transformation, and then visualize the entire patch at that location.
The patches in front are blended to those at the back, with patches on top have larger weights.
As the corresponding point clouds in 3D basically have the same appearance, for clarity we do not visualize the 3D point clouds in this result.

We investigate two scenarios: (1) visualizing descriptors of patches from different scenes, and then (2) descriptors of patches from the same scene.
For case (1), we demonstrate the t-SNE for $10,000$ image patches sampled from all scenes in 3DMatch~\cite{zeng20173dmatch}.
For case (2), we visualize the t-SNE of $10,000$ patches sampled in each scene in the SceneNN dataset~\cite{hua2016scenenn}.
As can be seen, in all embeddings, similar patches are clustered to each other.
It is worth noting that patches close to each other shares some similarity not only in colors but also in structures.
This demonstrates the effectiveness of the proposed network in learning representative and discriminative features.

\begin{figure*}[t]
    \centering
    \includegraphics[width=0.25\linewidth]{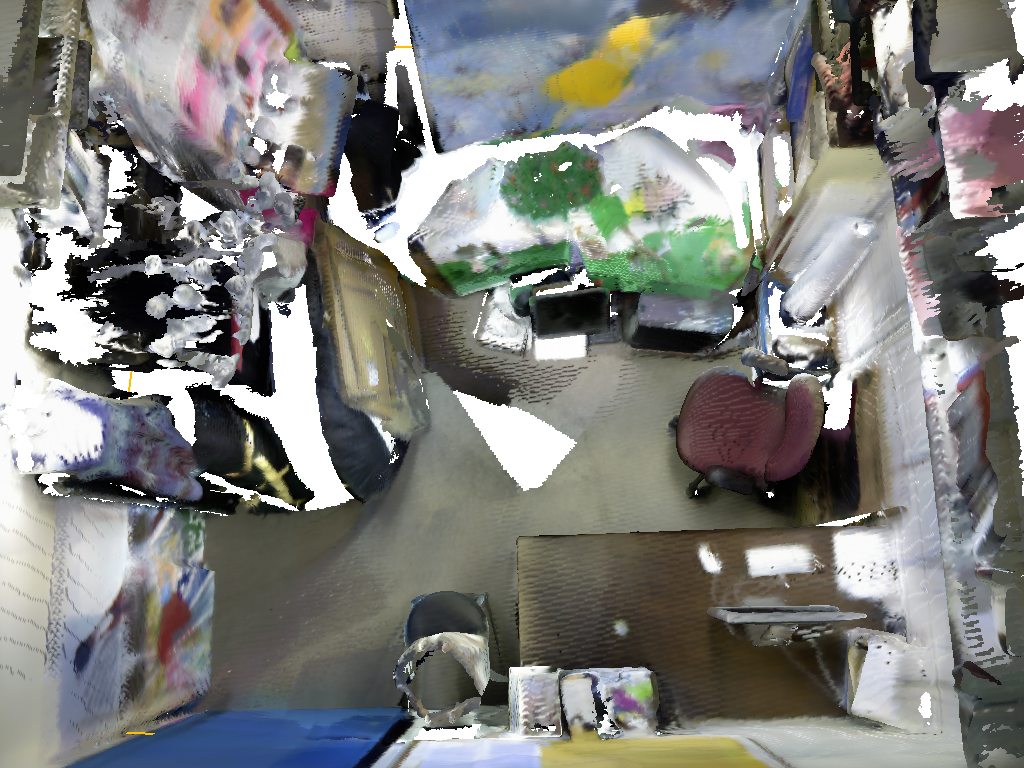}\\
    \includegraphics[width=\linewidth]{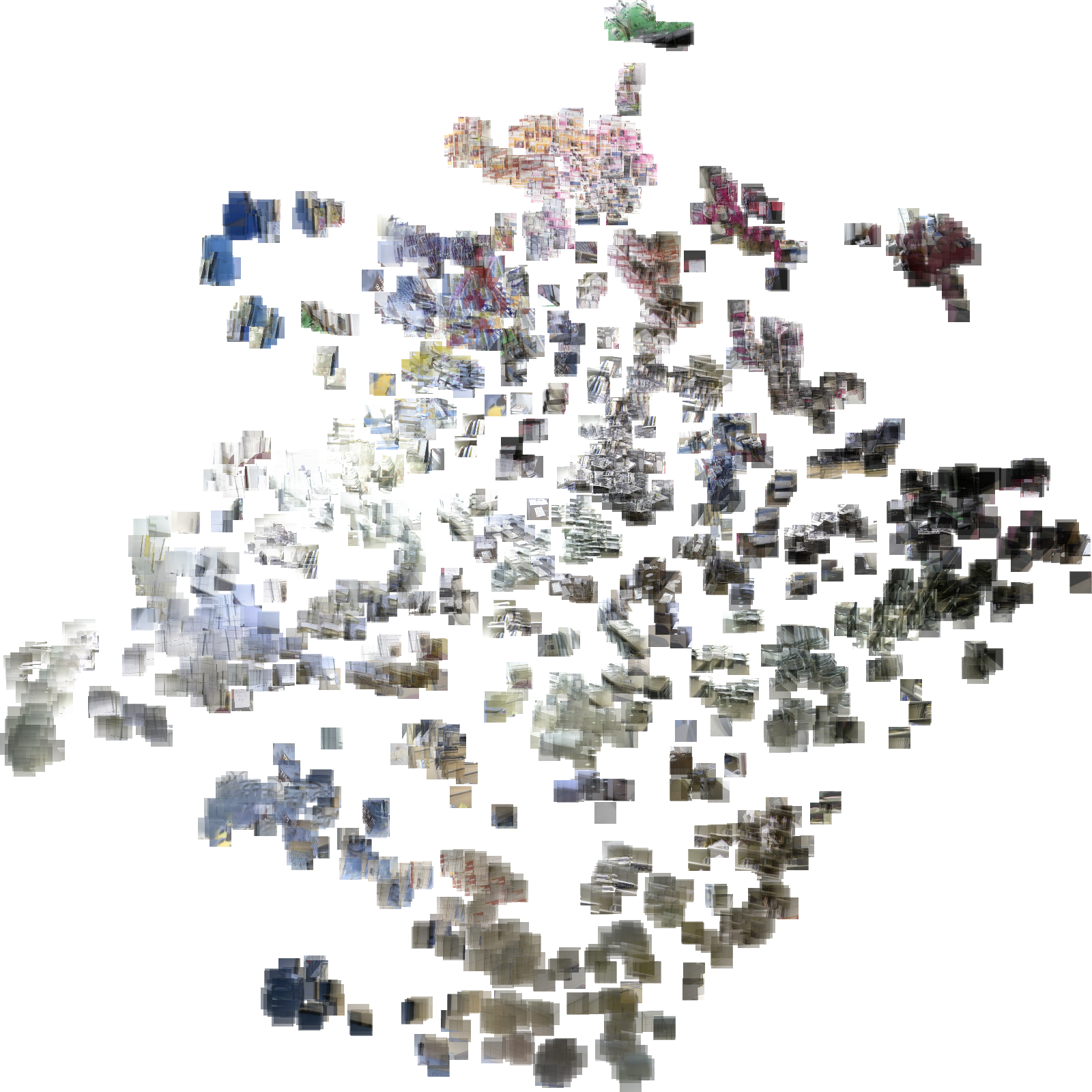}
    \caption{
      \textbf{t-SNE embedding of the descriptors in scene \texttt{073} (top view) in SceneNN dataset~\cite{hua2016scenenn}.}
      Best view in zoom and color.
    }
    \label{fig:tsne-scenenn}
\end{figure*}

\begin{figure*}[t]
    \centering
    \includegraphics[width=\linewidth]{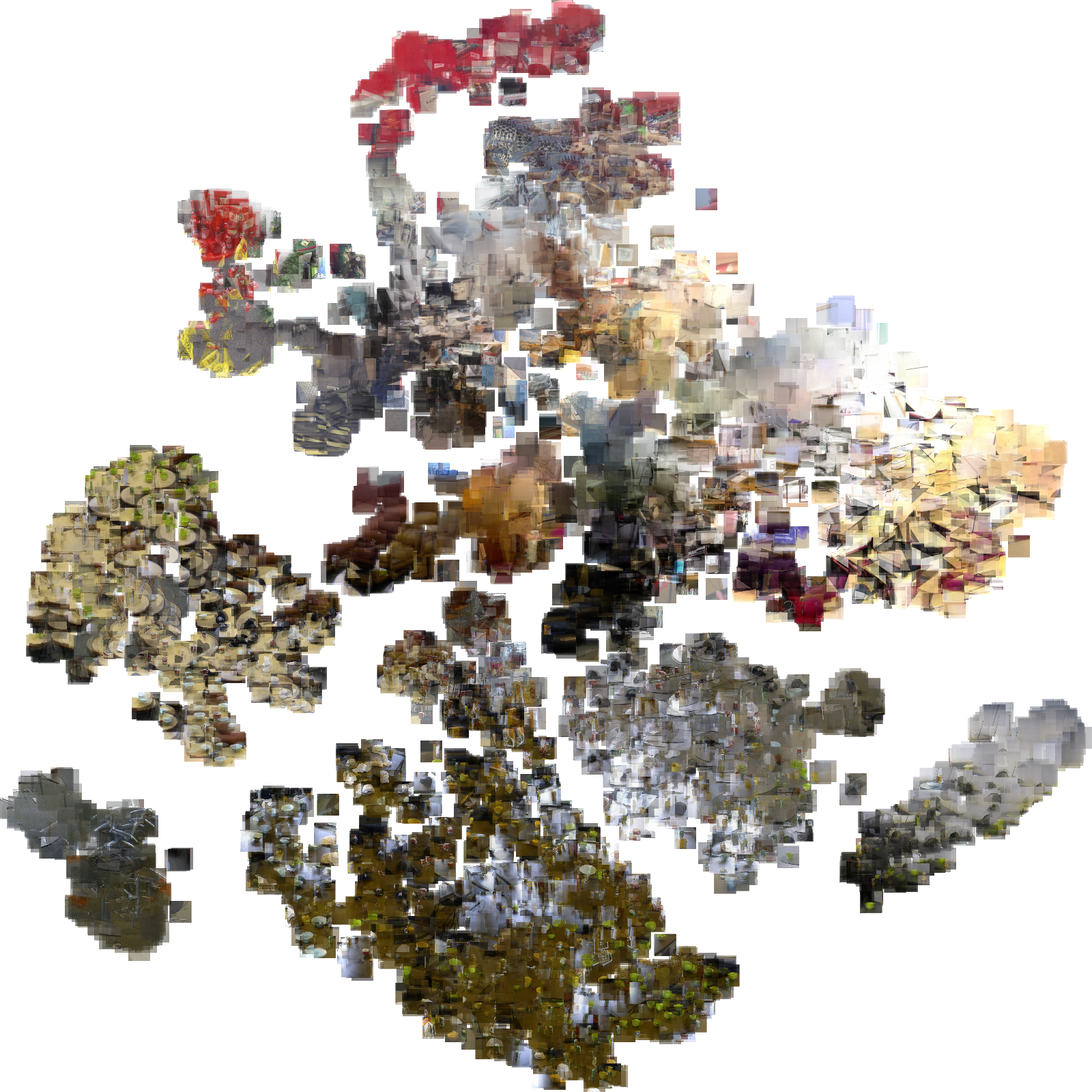}
    \caption{
      \textbf{t-SNE embedding of the descriptors in 3DMatch dataset~\cite{zeng20173dmatch}}.
      Best view in zoom and color.
    }
    \label{fig:tsne-3dmatch}
\end{figure*}

In addition, by comparing the complexity of t-SNE from case (1) and case (2), we found that the descriptors of patches in the same scene (Figure~\ref{fig:tsne-scenenn}) have less well-separated clusters than those sampled from different scenes (Figure~\ref{fig:tsne-3dmatch}).
This can be explained by the fact that objects a scene tend to be more correlated.
Moreover, objects in SceneNN are highly cluttered, causing a 3D region to often contain a few small objects, which renders the feature learning within a scene much more challenging.
Learning more robust features in this scenario would be an interesting future work.

{
  \small
  \bibliographystyle{aaai}
  \bibliography{ref}
}

\end{document}